%% file: icml_arxiv.tex
\documentclass{article}

\usepackage[usenames, dvipsnames]{color} 
\usepackage{microtype, graphicx, booktabs,verbatim, algorithm, algorithmic, amsmath, amssymb, caption, subcaption, bbm, mathrsfs, amsthm, tikz, url}
\usepackage[framemethod=TikZ]{mdframed}
\usepackage{natbib}

\usepackage{hyperref}

\usepackage[accepted]{icml2021}

\icmltitlerunning{Deciding What to Learn: A Rate-Distortion Approach}

\input{commands}

\def\environment{\mathcal{E}}
\def\environments{\Theta}

\def\actions{\mathcal{A}}

\def\observations{\mathcal{O}}

\def\E{\mathbb{E}}
\def\F{\mathbb{F}}
\def\Pr{\mathbb{P}}

\def\1{\mathbf{1}}

\newif\ifsubmit
\submitfalse
\ifsubmit
\newcommand{\dnote}[1]{}
\newcommand{\bnote}[1]{}
\else
\newcommand{\dnote}[1]{\textcolor{blue}{Dilip: #1}}
\newcommand{\bnote}[1]{\textcolor{orange}{Ben: #1}}
\fi

\begin{document}

\twocolumn[
\icmltitle{Deciding What to Learn: A Rate-Distortion Approach}



\icmlsetsymbol{equal}{*}

\begin{icmlauthorlist}
\icmlauthor{Dilip Arumugam}{st}
\icmlauthor{Benjamin Van Roy}{st}
\end{icmlauthorlist}

\icmlaffiliation{st}{Stanford University, California, USA}

\icmlcorrespondingauthor{Dilip Arumugam}{dilip@cs.stanford.edu}

\icmlkeywords{Sequential Decision-Making, Information Theory}

\vskip 0.3in
]



\printAffiliationsAndNotice{}  

\begin{abstract}
Agents that learn to select optimal actions represent a prominent focus of the sequential decision-making literature. In the face of a complex environment or constraints on time and resources, however, aiming to synthesize such an optimal policy can become infeasible. These scenarios give rise to an important trade-off between the information an agent must acquire to learn and the sub-optimality of the resulting policy. While an agent designer has a preference for how this trade-off is resolved, existing approaches further require that the designer translate these preferences into a fixed learning target for the agent. In this work, leveraging rate-distortion theory, we automate this process such that the designer need only express their preferences via a single hyperparameter and the agent is endowed with the ability to compute its own learning targets that best achieve the desired trade-off. We establish a general bound on expected discounted regret for an agent that decides what to learn in this manner along with computational experiments that illustrate the expressiveness of designer preferences and even show improvements over Thompson sampling in identifying an optimal policy.
\end{abstract}

\section{Introduction}
\label{sec:intro}

Learning is a process of acquiring information that reduces an agent's uncertainty about its environment.  Anything that an agent may endeavor to learn requires obtaining a precise amount of information about the environment; naturally, as measured by this requisite information, some things are easier to learn than others. When interacting with a complex environment, however, the agent is spoiled for choice as there is too much to learn within any reasonable time frame, and the agent must prioritize.  A simple approach is to designate a {\it learning target}, which can be thought of as a corpus of information that, while insufficient to fully identify the environment, suffices to guide effective decisions.  Then, the agent can prioritize gathering of information about this learning target.

One possible learning target, which has dominated the bandit-learning literature~\citep{bubeck2012regret,lattimore2020bandit}, is an action-selection policy that would be optimal given full information about the environment.  While suitable for simple environments, like multi-armed bandits with few arms, this concept does not scale well with the size of the action space. Moreover, in complex environments, there is typically too much to learn about the optimal policy within any reasonable time frame.

Recent work has highlighted conditions under which it is helpful to target a near-optimal or \textit{satisficing} policy~\citep{russo2018satisficing}. Such a learning target is not without precedent and has been studied implicitly in a variety of contexts~\citep{bubeck2011x,kleinberg2008multi,rusmevichientong2010linearly,ryzhov2012knowledge,deshpande2012linear,berry1997,Wang2008AlgorithmsFI,bonald2013two}. There is an important tension between information requirements for policy learning and policy performance; as one is more permissive of increasingly sub-optimal policies, the requisite amount of information for learning such policies decreases. Crucially, a satisficing policy can be manually specified by an agent designer in order to strike the desired balance. To do so, however, it is incumbent upon the designer to have sufficient knowledge of the problem structure in order to negotiate the information-performance trade-off.  

We consider the design of an agent that selects its own learning target.  This shifts the agent designer's role from specifying one to endowing the agent with the ability to designate and to suitably adapt the target as learning progresses. The designer can specify the general form of this learning target as part of the scaffold for a learning algorithm. More traditional, fixed-target learning algorithms can then be repurposed as subroutines an agent may use to achieve its own goals.  We introduce in this paper what is possibly the first principled approach to address a fundamental question: \textit{how should an agent decide what to learn?}

As a first step, this work offers one concrete answer to this question by introducing an agent that adaptively learns target actions. To endow this agent with the ability to reason about the information-performance trade-off autonomously, we employ rate-distortion theory~\citep{shannon1959coding,berger1971rate}, building on connections to sequential decision-making made by \citet{russo2018satisficing}. With an appropriately chosen distortion measure, the canonical rate-distortion function precisely characterizes the trade-off between the information required for policy learning and policy performance. Rather than placing the burden on the agent designer to procure the solution to a single rate-distortion function on behalf the agent, we instead place the onus upon the agent to solve a rate-distortion function in each time period and gradually adapt its self-designated target action. We recognize that computation of rate-distortion functions is a well-studied problem of the information theory community for which an elegant solution already exists as the classic Blahut-Arimoto algorithm~\citep{blahut1972computation,arimoto1972algorithm}. Accordingly, we begin by introducing a variant of Thompson sampling which uses the Blahut-Arimoto algorithm as a subroutine for computing a target action in each time period that achieves the rate-distortion limit. We then prove a bound on the expected discounted regret for this algorithm, differing from previous information-theoretic analyses in its treatment of a learning target that changes in each time period. Finally, we conclude with a series of computational experiments that highlight the efficacy of our procedure in enabling an agent to target desired points along the information-performance trade-off curve.

The paper proceeds as follows: in Section \ref{sec:back} we briefly discuss background material before clarifying the connections between our approach and rate-distortion theory in Section \ref{sec:dm_rdt}. Due to space constraints, we relegate an overview of prior work to the appendix. We introduce our main algorithm in Section \ref{sec:appr} before finally presenting a corresponding regret analysis and supporting computational experiments in Sections \ref{sec:regret} and \ref{sec:exps}, respectively.

\section{Background}
\label{sec:back}

In this section, we begin with an overview of several standard quantities in information theory. For more background on information theory, see \citet{cover2012elements}. We conclude the section with a brief outline of rate-distortion theory.

\subsection{Information Theory}
\label{sec:back_it}

Consider three random variables $X, Y, Z$ defined on a probability space $(\Omega,\bF,\bP)$. We define entropy, conditional entropy, mutual information, and conditional mutual information as follows:
\begin{align*}
    \bH(X) &= -\bE[\log(\bP(X \in \cdot ))] \\
    \bH(Y|X) &= -\bE[\log(\bP(Y \in \cdot |X))] \\
    \bI(X;Y) &= \bH(X) - \bH(X | Y) = \bH(Y) - \bH(Y | X) \\
    \bI(X;Y|Z) &= \bH(X|Z) - \bH(X | Y,Z) = \bH(Y|Z) - \bH(Y | X,Z) \\
\end{align*}

Importantly, the multivariate mutual information between a single random variable $X$ and another sequence of random variables $Z_1, \ldots Z_n$ decomposes via the chain rule of mutual information:

\begin{align*}
    \bI(X;Z_1,\ldots,Z_n) &= \sum\limits_{i=1}^n \bI(X;Z_i|Z_1,\ldots,Z_{i-1})
\end{align*}

\subsection{Rate-Distortion Theory}
\label{sec:back_rdt}

Rate-distortion theory is a sub-area of information theory concerned with lossy compression and the achievability of coding schemes that maximally compress while adhering to a desired upper bound on error or loss of fidelity~\citep{shannon1959coding,berger1971rate,cover2012elements}. More formally, consider a random variable $X$ with fixed distribution $p(x) = \bP(X = x)$ that represents an information source along with a random variable $\hat{X}$ that corresponds to a channel output. Given a distortion measure $d:\mc{X} \times \hat{\mc{X}} \mapsto \bR_{\geq 0}$ and a desired upper bound on distortion $D$, the rate-distortion function is defined as:
\begin{align}
    \mc{R}(D) &= \inf\limits_{\hat{X} \in \Lambda} \mc{I}(X;\hat{X})
    \label{eqn:rd_constr_obj}
\end{align}
quantifying the minimum number of bits (on average) that must be communicated from $X$ across a channel in order to adhere to the specified expected distortion threshold $D$. Here, the infimum is taken over $\Lambda = \{\hat{X} : \bE\left[d(X,\hat{X})\right] \leq D \}$ representing the set of all random variables $\hat{X}: \Omega \mapsto \hat{\mc{X}}$ which which satisfy the constraint on expected distortion. Intuitively, a higher rate corresponds to requiring more bits of information and smaller information loss between $X$ and $\hat{X}$, enabling higher-fidelity reconstruction (lower distortion); conversely, lower rates reflect more substantial information loss, potentially exceeding the tolerance on distortion $D$. 

\begin{fact}
$\mc{R}(D)$ is a non-negative, convex, and  monotonically-decreasing function in $D$~\citep{cover2012elements}.
\label{fact:rd_convex}
\end{fact}

Some readers may be more familiar with the related problem of computing channel capacity; while the rate-distortion function considers a fixed information source $p(x)$ and optimizes for a channel $p(\hat{x}|x) = \bP(\hat{X} = \hat{x} | X = x)$ that minimizes distortion, the channel-capacity function considers a fixed channel and optimizes for the information source that maximizes throughput.

\section{Sequential Decision-Making \& Rate-Distortion Theory}
\label{sec:dm_rdt}

\subsection{Problem Formulation}
\label{sec:prblm}

We define all random variables with respect to a common probability space $(\Omega, \F, \Pr)$; all events are determined by a random outcome $\omega \in \Omega$.  An agent interacts with an unknown environment $\environment$, which is itself a random variable.  The interaction generates a history $H_t = (A_0, O_1, A_1, O_2, \ldots, O_t)$ of actions and observations that take values in finite sets $\actions$ and $\observations$.  Initial uncertainty about the environment is reflected by probabilities $\Pr(\environment \in \cdot)$ where $\environment$ has support on $\Theta$ and, as the history unfolds, what can be learned is represented by conditional probabilities $\Pr(\environment \in \cdot | H_t)$.

Actions are independent of the environment conditioned on history, $A_{t+1} \perp \environment | H_t$.  This reflects the fact that the agent selects actions based only on history and, possibly, algorithmic randomness.  It may be helpful to think of the actions as being selected by an {\it admissible policy} $\pi(a | H_t) = \Pr(A_t = a| H_t)$, which assigns a probability to each action $a \in \actions$ given the history.  By {\it admissible}, we mean that action probabilities are determined by history and do not depend on further information about the environment.

We assume that observations are independent of history conditioned on the environment and most recent action, $O_{t+1} \perp H_t | (\environment, A_t)$.  Note that this precludes delayed consequences, and we will restrict attention in this paper to such environments.  Further, we assume a stationary environment such that conditional observation probabilities $\Pr(O_{t+1} | \environment, A_t)$ do not depend on $t$.

Upon each observation, the agent enjoys a reward $R_{t+1} = r(A_t, O_{t+1})$ where $r: \mc{A} \times \mc{O} \mapsto \bR$ is a deterministic function. Let $\overline{r}(a) = \E[R_{t+1} | A_t = a, \environment]$ denote mean reward and note that $\overline{r}$ is itself a random variable since it depends on $\environment$.  Let $A_\star$ be an action that maximizes the expected mean reward $\E[\overline{r}(A_\star)]$ and let $R_\star = \overline{r}(A_\star)$.  Note that $A_\star$ and $R_\star$ are random variables, as they depend on $\environment$.  It may be helpful to think of $A_\star$ as generated by an optimal policy $\pi_\star(a) = \Pr(A_t = a | \environment)$, which is {\it inadmissible}, in the sense that it depends on the environment, not just the history. Traditionally, the performance of an admissible policy $\pi$ at any time period $\tau = 0, 1, 2,\ldots$ is quantified by its regret: $$\bE\left[\sum\limits_{t=\tau}^\infty R_\star - R_{t+1} \Big| H_\tau \right].$$

While this is a suitable measure of asymptotic performance, we follow suit with \citet{russo2018satisficing} and examine expected discounted regret $$\bE\left[\sum\limits_{t=\tau}^\infty \gamma^{t-\tau} (R_\star - R_{t+1}) \Big| H_\tau \right],$$ where the discount factor $\gamma \in [0,1)$ helps regulate the agent's preference for minimizing near-term versus long-term performance shortfall. 

\subsection{Target Actions}

In the course of identifying an optimal policy, we take $\mathbb{H}(A_\star)$ to denote the bits of information an agent must acquire in order to identify $A_\star$. \citet{russo2016information} offer a novel information-theoretic analysis of Thompson sampling~\citep{thompson1933likelihood} whose corresponding regret bound depends on $\mathbb{H}(A_\star)$. Due to the non-negativity of conditional entropy, $\mathbb{H}(A_\star | \environment) \geq 0$, it follows that the entropy of $A_\star$ upper bounds the mutual information between $A_\star$ and $\environment$, $\mathbb{H}(A_\star) \geq \mathbb{H}(A_\star) - \mathbb{H}(A_\star | \environment) = \mathbb{I}(A_\star; \environment)$, which is tight when the optimal action $A_\star$ is a deterministic function of $\environment$.

When faced with a complex environment $\environment$, acquiring these $\bH(A_\star)$ bits of information for optimal behavior may be exceptionally difficult. While Thompson sampling is a simple yet effective algorithm with widespread empirical success in synthesizing optimal policies~\citep{chapelle2011empirical,russo2018tutorial}, it can fall short in these more challenging learning settings. \citet{russo2018satisficing} first drew awareness to this issue, highlighting several examples where Thompson sampling struggles in the face of a large, possibly infinite, action set or a time-sensitivity constraint on learning. In short, the problem stems from the fact that Thompson sampling will select new, untested actions in each time period, rapidly becoming inefficient as the number of actions grows.

\citet{russo2018satisficing} introduce the notion of satisficing actions $\tilde{A}$, in lieu of optimal actions, as a remedy to the aforementioned issues. The core premise of this alternative learning target is that a deliberately sub-optimal action should require the agent to learn fewer bits of information about the environment in order to identify a corresponding satisficing policy. Their proposed satisficing Thompson sampling algorithm makes the natural modification of probability matching with respect to the agent's posterior beliefs over $\tilde{A}$, given the current history, such that $A_t \sim \bP(\tilde{A} = \cdot | H_t)$. Crucially, \citet{russo2018satisficing} draw an interesting connection between the specification of satisficing actions and rate-distortion theory. Taking the distortion function to be the instantaneous expected regret conditioned on a realization of the environment, $d(\tilde{a},e) = \bE[\overline{r}(A_\star) - \overline{r}(a) | \environment = e]$, they study the corresponding rate-distortion function 
\begin{align}
    \mc{R}(D) = \inf\limits_{\tilde{A} \in \tilde{\mc{A}}} \mathbb{I}(\tilde{A};\environment)
    \label{eqn:rd_satact}
\end{align} 
where $\tilde{\mc{A}} = \{\tilde{A} : \expect{d(\tilde{A},\environment)} \leq D, \tilde{A} \perp H_t | \environment, \forall t\}$ denotes the set of all random variables $\tilde{A}: \Omega \mapsto \mc{A}$ that are conditionally-independent from all histories given the environment $\environment$ and adhere to the distortion constraint.
Applying Fact \ref{fact:rd_convex}, we immediately recover the following:
\begin{fact}
For any $D > 0$, $\mathbb{H}(A_\star) \geq \mathbb{H}(A_\star) - \mathbb{H}(A_\star | \environment) = \mathbb{I}(A_\star; \environment) = \mc{R}(0) \geq \mc{R}(D) = \bI(\tilde{A};\environment)$
\label{fact:sa_less_info}
\end{fact}
\noindent
which confirms a crucial desideratum for satisficing actions; namely, that an agent must acquire fewer bits of information about $\environment$ in order to learn a satisficing action, relative to learning an optimal action. Moreover, following an analogue of the information-theoretic analysis of \citet{russo2016information}, \citet{russo2018satisficing} prove an information-theoretic regret bound that depends on the value of the rate-distortion function, rather than the entropy. While this performance guarantee highlights an interesting and useful link between sequential decision-making and rate-distortion theory, there is no guarantee that a manually-specified satisficing action $\tilde{A}$ will achieve the rate-distortion limit as desired. Thus, an agent that can manufacture its own satisficing actions which achieve the rate-distortion limit stands to dramatically outperform any hand-crafted $\tilde{A}$. To make the distinction between the manually-specified satisficing actions of prior work, we use the term \textit{target actions} to refer to the agent's self-designated learning targets which explicitly differ from satisficing actions in that the are (1) computed by the agent, (2) adapted over time according the agent's current knowledge of the environment $\environment$, and (3) achieve the rate-distortion limit in each time period.

Agents we consider can forgo the aim of learning an optimal action and instead try to learn a {\it target action}.  Formally, a {\it target action} $\tilde{A}$ is a random variable that be thought of as generated by an inadmissible policy $\tilde{\pi}(a) = \Pr(\tilde{A} = a | \environment)$.  Similarly with $A_\star$, a target action may depend on the environment, not just the history. Moreover, a target action is a random variable $\tilde{A}$ that satisfies $H_t \perp \tilde{A} | \environment$ for all $t$.  In other words, observations do not provide information about $\tilde{A}$ beyond what the environment would. As it based upon an inadmissible policy, a target action can change along with the agent's beliefs over the environment $\bP(\environment \in \cdot | H_t)$. This represents another key distinction between target actions that an agent can modify to reflect its updated knowledge about the environment and manually-specified satisficing actions that act as a fixed learning objective (much like optimal actions $A_\star$). We use $\tilde{A}_t$ to denote the target action computed in time period $t$ according to the distortion function $d(a,e|H_t) = \E[(\overline{r}(A_\star) - \overline{r}(a))^2 | \environment = e, H_t])$. Consequently, this induces a sequence of rate-distortion functions, one for each time period, each of which is conditioned on the agent's history $H_t$. In the next section, we discuss a classic approach for computing a single, arbitrary rate-distortion function before introducing a variant of Thompson sampling that applies this method to compute target actions in each time period.

\section{Approach}
\label{sec:appr}

\subsection{Notation}

At various points going forward, it will be necessary to refer to the mutual information between two random variables conditioned upon a specific realization of an agent's history at some time period $t$. For convenience, we will denote this as $$\bI_t(X;Y) = \bI(X;Y|H_t = H_t).$$ This notation will also apply analogously to the conditional mutual information $$\bI_t(X;Y|Z) = \bI(X;Y|H_t = H_t, Z).$$ Note that their dependence on the realization of random history $H_t$ makes both $\bI_t(X;Y)$ and $\bI_t(X;Y|Z)$ random variables themselves. The traditional notion of conditional mutual information which uses the random variable $H_t$ arises by integrating over this randomness:
\begin{align*}
    \bE\left[\bI_t(X;Y)\right] &= \bI(X;Y|H_t) \\
    \bE\left[\bI_t(X;Y|Z)\right] &= \bI(X;Y|H_t,Z)
\end{align*}

Additionally, we will also adopt a similar notation to express a conditional expectation given the random history $H_t$: $$\bE_t\left[X\right] = \bE\left[X|H_t\right].$$

\subsection{Blahut-Arimoto Satisficing Thompson Sampling}
\label{sec:baa}

A classic algorithm for carrying out the constrained optimization problem captured in the rate-distortion function is the Blahut-Arimoto algorithm~\citep{blahut1972computation,arimoto1972algorithm}. While the first step in the derivation of the algorithm is to start with the Lagrangian of the constrained objective, we will adopt a different notation to recognize the sequence of rate-distortion functions an agent must solve as its history expands. Namely, consider a loss function that, given history $H_t$, assesses a target action:
$$\mathcal{L}_\beta(\tilde{A}|H_t) = \mathbb{I}_t(\environment; \tilde{A}) +  \beta \E_t\left[(\overline{r}(A_\star) - \overline{r}(\tilde{A}))^2\right].$$
The first term can be interpreted as the number of bits of information from the environment required to identify target action, which we refer to as the {\it information rate} of $\tilde{A}$.  The second term is a measure of distortion -- the expected squared error between mean rewards generated by the target action versus an optimal action -- scaled by a constant $\beta \in \bR_{\geq 0}$ representing a Lagrange multiplier.  Hence, this loss-function captures a rate-distortion trade-off.  An optimal action minimizes distortion but, via Fact \ref{fact:sa_less_info}, may require a high rate.  An uninformed action has a rate of zero but results in high distortion. Our goal is for an agent designer to use the $\beta$ hyperparameter to express a preference for the ease of learning versus the tolerable level of sub-optimality whereas it is the agent's responsibility to identify the appropriate target action $\tilde{A}_t$ that best reflects these preferences~\citep{singh2010separating}.

The Blahut-Arimoto Algorithm can be applied to identify a target action $\tilde{A}$ that minimizes this loss function.  The algorithm is initialized with environment-dependent target action probabilities $\tilde{p}_0$, and generates a sequence of iterates $\tilde{p}_1,\tilde{p}_2, \ldots$, converging on probabilities $\tilde{p}_\star$ such that $\tilde{p}_\star(a|e) = \Pr(\tilde{A} = a|\environment = e)$ for all $a \in \actions$ and $e \in \Theta$.  Each iteration carries out two steps.  The first computes marginal probabilities of the target action
$$\tilde{q}_k(a) = \E_t[\tilde{p}_k(a|\environment)] \qquad \forall a \in \actions,$$
while the second updates environment-dependent target action probabilities, $\forall a \in \actions, e \in \environments,$
$$\tilde{p}_{k+1}(a |e) = \tfrac{\tilde{q}_k(a) \exp(-\beta \E_t[(\overline{r}(A_\star) - \overline{r}(a))^2 | \environment = e])}{\sum_{a' \in \actions} \tilde{q}_k(a') \exp(-\beta \E_t[(\overline{r}(A_\star) - \overline{r}(a'))^2 | \environment = e])}.$$

A standard choice for the initial channel parameters $\tilde{p}_0(a|e)$ is the uniform distribution. Again, $\beta$ now subsumes the role of $D$ in Equation \ref{eqn:rd_satact} for expressing the desired prioritization of minimizing rate (lower $\bI(\tilde{A}_t;\environment)$) versus minimizing distortion (lower  $d(a,e|H_t) = \E_t[(\overline{r}(A_\star) - \overline{r}(a))^2 | \environment = e])$). Notice that as $\beta \ra \infty$, $\tilde{p}_{k+1}(a |e)$ sharpens to a max, placing all probability mass on the realization of $\tilde{A}$ that minimizes distortion; consequently, $\tilde{p}_\star(a|e) = \bP(\tilde{A} = a | \environment = e) = \bP(A_\star = a | \environment = e)$ and we recover the standard learning target of Thompson sampling.

Just as Thompson sampling selects actions according to the probability of being optimal $\bP(A_t = a | H_{t-1}) = \bP(A^\star = a | H_{t-1})$, our BLahut-Arimoto Satisficing Thompson Sampling (BLASTS) algorithm selects actions according to their probability of being the target action $\tilde{A}_t$ that achieves the rate-distortion limit. We present the BLASTS algorithm as Algorithm \ref{alg:blasts}.


\section{Regret Analysis}
\label{sec:regret}

Abstracting away the precise details of BLASTS, we can consider a coarsely-defined algorithm that selects each action $A_t$ as follows: \textbf{(1)} identify a target action $\tilde{A}_t$ that minimizes a loss function $\mathcal{L}_\beta(\cdot|H_t)$ and \textbf{(2)} sample $A_t \sim \Pr(\tilde{A}_t = \cdot | H_t)$. Recall that the loss function is defined, for any target action $\tilde{A}$, by
$$\mathcal{L}_\beta(\tilde{A}|H_t) = \mathbb{I}_t(\environment; \tilde{A}) +  \beta \E_t\left[(\overline{r}(A_\star) - \overline{r}(\tilde{A}))^2 \right].$$
Due to space constraints, the proofs associated with all of the following results can be found in the appendix. The following result helps establish that the expected loss of any target action decreases as observations accumulate.

\input{algorithms/blasts_alg_no_comments}

\begin{lemma}
\label{le:information-supermartingale}
For all $\beta > 0$, target actions $\tilde{A}$, and $t = 0,1,2,\ldots$, 
$$\E_t[\mathcal{L}_\beta(\tilde{A} | H_{t+1})] = \mathcal{L}_\beta(\tilde{A} | H_t) - \mathbb{I}_t(\tilde{A};(A_t, O_{t+1})).$$
\end{lemma}
As a consequence of the above, the following lemma assures that expected loss decreases as target actions are adapted.  It also suggests that there are two sources of decrease in loss: (1) a possible decrease in shifting from target $\tilde{A}_t$ to $\tilde{A}_{t+1}$ and (2) a decrease of $\mathbb{I}_t(\tilde{A}_t; (A_t, O_{t+1}))$ from observing the interaction $(A_t, O_{t+1})$.  The former reflects the agent's improved ability to select a suitable target, and the latter captures information gained about the previous target.  The proof of the lemma follows immediately from Lemma \ref{le:information-supermartingale} and the fact that $\tilde{A}_{t+1}$ minimizes $\mathcal{L}_\beta(\tilde{A}_{t+1} | H_{t+1})$, by definition.
\begin{lemma}
For all $\beta > 0$, target actions $\tilde{A}$, and $t = 0,1,2,\ldots$, 
$$\E[\mathcal{L}_\beta(\tilde{A}_{t+1} | H_{t+1}) | H_t] \leq \mathcal{L}_\beta(\tilde{A}_t | H_t) - \mathbb{I}_t(\tilde{A}_t;(A_t, O_{t+1})).$$
\label{le:adapt_targets}
\end{lemma}
Note that, for all $t$, loss is non-negative and bounded by mutual information between the optimal action and the environment (since optimal actions incur a distortion of 0):
$$\mathcal{L}_\beta(\tilde{A}_t | H_t) \leq \mathcal{L}_\beta(A_\star | H_t) = \mathbb{I}_t(\environment; A_\star).$$
We therefore have the following corollary.
\begin{corollary}
For all $\beta > 0$ and $\tau = 0,1,2,\ldots$, 
$$\E\left[\sum_{t=\tau}^\infty \mathbb{I}_t(\tilde{A}_t; (A_t, O_{t+1})) \Big| H_\tau\right] 
\leq \mathbb{I}_\tau(\environment; A_\star).$$
\end{corollary}
The proof of Corollary 1 follows directly by applying the preceding inequality to the following  generalization that applies to any target action.
\begin{corollary}
\label{co:cumulative-information}
For all $\beta > 0$, target actions $\tilde{A}$, and $\tau = 0,1,2,\ldots$, 
$$\E_\tau\left[\sum_{t=\tau}^\infty \mathbb{I}_t(\tilde{A}_t; (A_t, O_{t+1})) \right] 
\leq \mathcal{L}_\beta(\tilde{A} | H_\tau).$$
\end{corollary}

Let $\Gamma$ be a constant such that 
$$\Gamma \geq \frac{\E_t[\overline{r}(\tilde{A}) - \overline{r}(A)]^2}{\mathbb{I}_t(\tilde{A}; A, O)},$$
for all histories $H_t$, target actions $\tilde{A}$, if the executed action $A$ is an independent sample drawn from the marginal distribution of $\tilde{A}$, and $O$ is the resulting observation. Thus, $\Gamma$ is an upper bound on the information ratio~\citep{russo2014learning,russo2016information,russo2018learning} for which existing information-theoretic analyses of worst-case finite-arm bandits and linear bandits provide explicit values of $\Gamma$ that satisfy this condition.

We can now establish our main results.  We omit the proof of Theorem \ref{thm:rb_optimact} as it is a special case of our subsequent result.
\begin{theorem}
If $\beta = \frac{1-\gamma^2}{(1-\gamma)^2 \Gamma}$ then, for all $\tau = 0,1,2,\ldots$,
$$\E_\tau\left[\sum_{t=\tau}^\infty \gamma^{t-\tau} (\overline{r}(A_\star) - \overline{r}(A_t)) \right] \leq 2 \sqrt{\frac{\Gamma \mathbb{I}_\tau(\environment; A_\star)}{1-\gamma^2}}.$$
\label{thm:rb_optimact}
\end{theorem}

In a complex environment with many actions, $\mathbb{I}(\environment; A_\star)$ can be extremely large, rendering the above result somewhat vacuous under such circumstances.  The next result offers a generalization, establishing a regret bound that can depend on the information content of any target action, including of course those that are much simpler than $A_\star$.
\begin{theorem}
If $\beta = \frac{1-\gamma^2}{(1-\gamma)^2 \Gamma}$ then, for all target actions $\tilde{A}$ and $\tau = 0,1,2,\ldots$,
\begin{align*}
    \E_\tau\left[\sum_{t=\tau}^\infty \gamma^{t-\tau} (\overline{r}(A_\star) - \overline{r}(A_t))\right] &\leq 2 \sqrt{\frac{\Gamma \mathbb{I}_\tau(\environment; \tilde{A})}{1-\gamma^2}} + \frac{2\epsilon}{1-\gamma},
\end{align*}
where $\epsilon = \sqrt{\E_\tau[(\overline{r}(A_\star) - \overline{r}(\tilde{A})^2]}$.
\label{thm:rb_targetact}
\end{theorem}

For the sake of completeness, we may derive the analogues of Corollary \ref{co:cumulative-information} and Theorem \ref{thm:rb_targetact} for the more traditional finite-horizon, undiscounted regret setting.

\begin{corollary}
\label{co:cumulative-information_undisc}
For all $\beta > 0$, target actions $\tilde{A}$, and $\tau = 0,1,2,\ldots$, 
$$\E_\tau\left[\sum_{t=\tau}^{T + \tau} \mathbb{I}_t(\tilde{A}_t; (A_t, O_{t+1})) \right] 
\leq \mathcal{L}_\beta(\tilde{A} | H_\tau).$$
\end{corollary}

\begin{theorem}
If $\beta = \frac{T}{\Gamma}$ then, for all target actions $\tilde{A}$ and $\tau = 0,1,2,\ldots$,
$$\E_\tau\left[\sum_{t=\tau}^{T + \tau} \overline{r}(A_\star) - \overline{r}(A_t)\right] \leq 2 \sqrt{\Gamma T \mathbb{I}_\tau(\environment; \tilde{A})} + 2T\epsilon,$$
where $\epsilon = \sqrt{\E[(\overline{r}(A_\star) - \overline{r}(\tilde{A})^2 | H_\tau]}$.
\label{thm:rb_targetact_undisc}
\end{theorem}

Notably, the information-theoretic regret bounds of Theorems \ref{thm:rb_targetact} and \ref{thm:rb_targetact_undisc} align with that of \citep{russo2018satisficing} as a sum of the difficulty associated with learning $\tilde{A}$ and the associated performance shortfall between $\tilde{A}$ and $A_\star$.

\section{Experiments}
\label{sec:exps}

\begin{figure}
\centering
\begin{subfigure}{.5\textwidth}
  \centering
  \includegraphics[width=.8\linewidth]{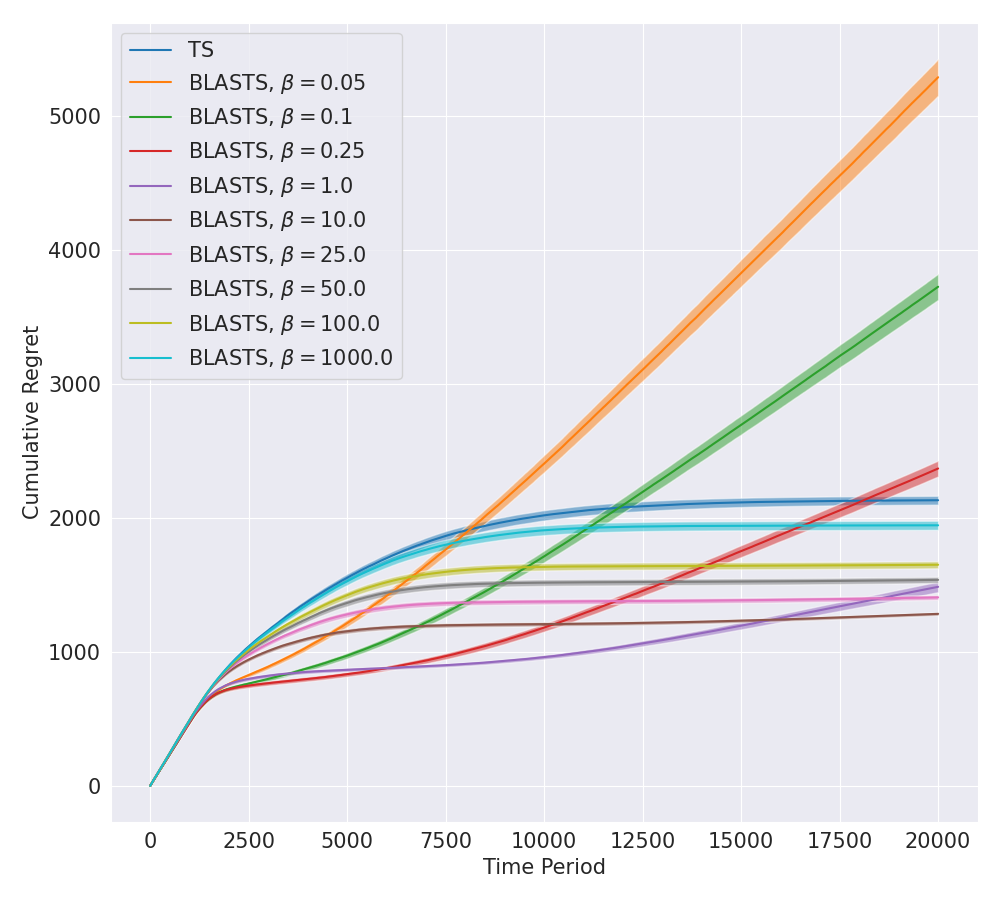}
  \caption{50 arms}
\end{subfigure}
\begin{subfigure}{.5\textwidth}
  \centering
  \includegraphics[width=.8\linewidth]{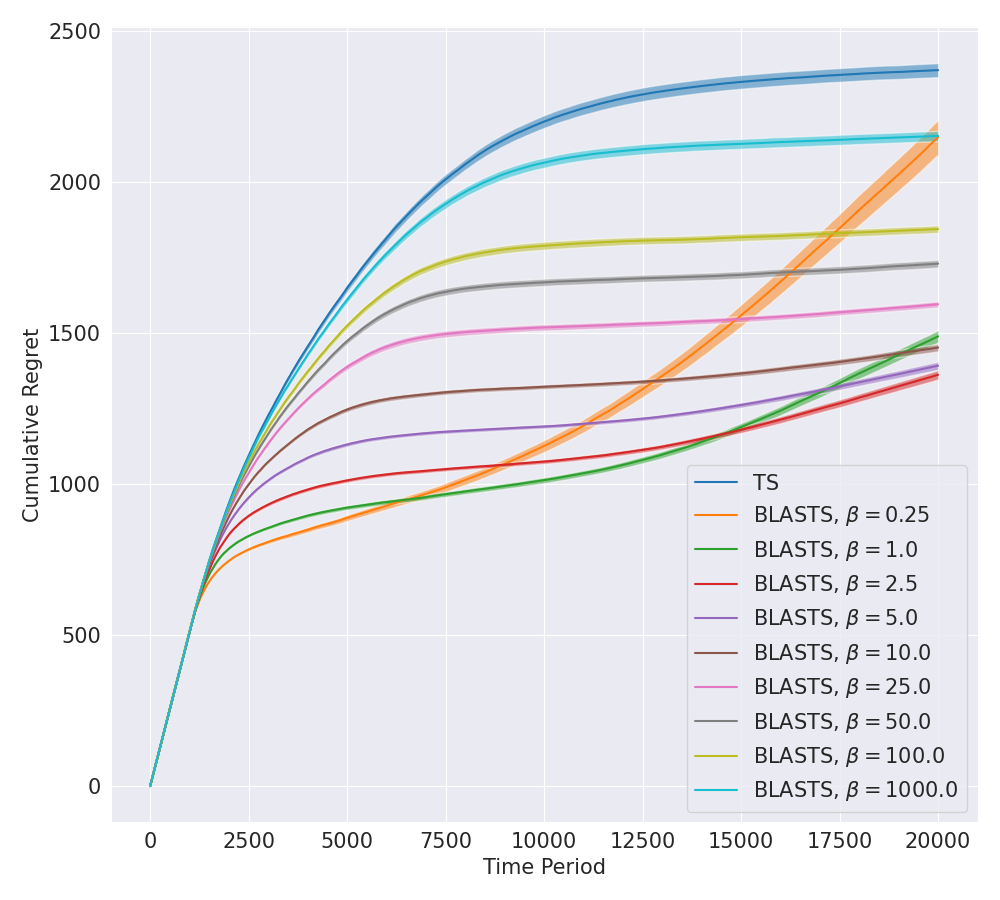}
  \caption{250 arms}
\end{subfigure}
\caption{Bernoulli bandit with independent arms}
\label{fig:blasts_bern}
\end{figure}

In this section, we outline two sets of computational experiments that evaluate BLASTS against traditional Thompson sampling (TS). The primary goal of our experiments is to illustrate how BLASTS enables an agent to navigate the information-performance trade-off through the specification of $\beta$. To this end, we examine two commonly-studied multi-armed bandit problems and sweep across several values of $\beta$, benchmarking performance relative to Thompson sampling. In the course of doing so, we find that both settings offer a range of $\beta$ values which allow the agent to converge on the optimal policy with greater efficiency than Thompson sampling.

In all of our experiments, we use linear hypermodels~\citep{dwaracherla2020hypermodels} as a common choice for representing an agent's epistemic uncertainty over the environment $\environment$. While several prior works have made use of finite ensembles for representing an agent's posterior beliefs over environment parameters~\citep{osband2016deep,lu2017ensemble}, hypermodels offer a more computationally-tractable approach that demonstrably scales better with a large number of actions. For an independent multi-armed bandit problem with $K$ actions, a linear hypermodel takes as input an index sample $z \sim \mc{N}(0,I_K)$ and computes a single posterior sample as $f_\nu(z) = \mu + \sigma z$ where the parameters $\nu = (\mu \in \bR^K, \sigma \in \bR^K)$ are incrementally updated via gradient descent to minimize a bootstrapped loss function. Due to space constraints, we refer readers to \citep{dwaracherla2020hypermodels} for the precise details of this loss function and further information about hypermodels. It is important to note that both Thompson sampling and BLASTS are agnostic to this modeling choice and are compatible with any approach for representing an agent's uncertainty about the environment. We use a noise variance of 0.1, a prior variance of 1.0, and a batch size of 1024 throughout all experiments while using Adam~\citep{kingma2014adam} to optimize hypermodel parameters with a learning rate of 0.001. 

We leverage an existing implementation of the Blahut-Arimoto algorithm for all experiments~\citep{dit}. The number of posterior samples used was fixed to 64 and the maximum number of iterations was set to 100, stopping early if the average distortion between two consecutive iterations fell below a small threshold. In preliminary experiments, we found better numerical stability when running the Blahut-Arimoto algorithm in base 2, rather than base $e$. To benchmark performance, we plot the (undiscounted) cumulative regret in each time period with shading to represent 95\% confidence intervals computed across 10 random seeds. 

\subsection{Independent Bernoulli \& Gaussian Bandits}

\begin{figure}[H]
\centering
\begin{subfigure}{.5\textwidth}
  \centering
  \includegraphics[width=.8\linewidth]{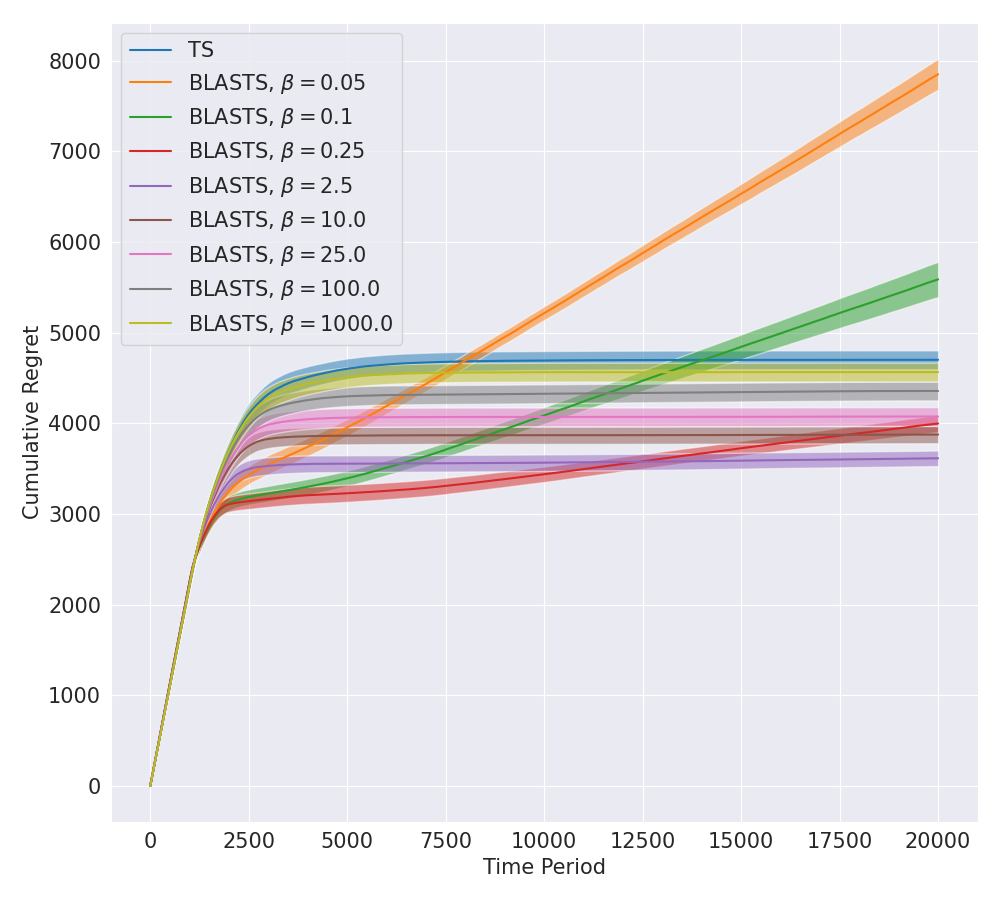}
  \caption{50 arms}
\end{subfigure}
\begin{subfigure}{.5\textwidth}
  \centering
  \includegraphics[width=.8\linewidth]{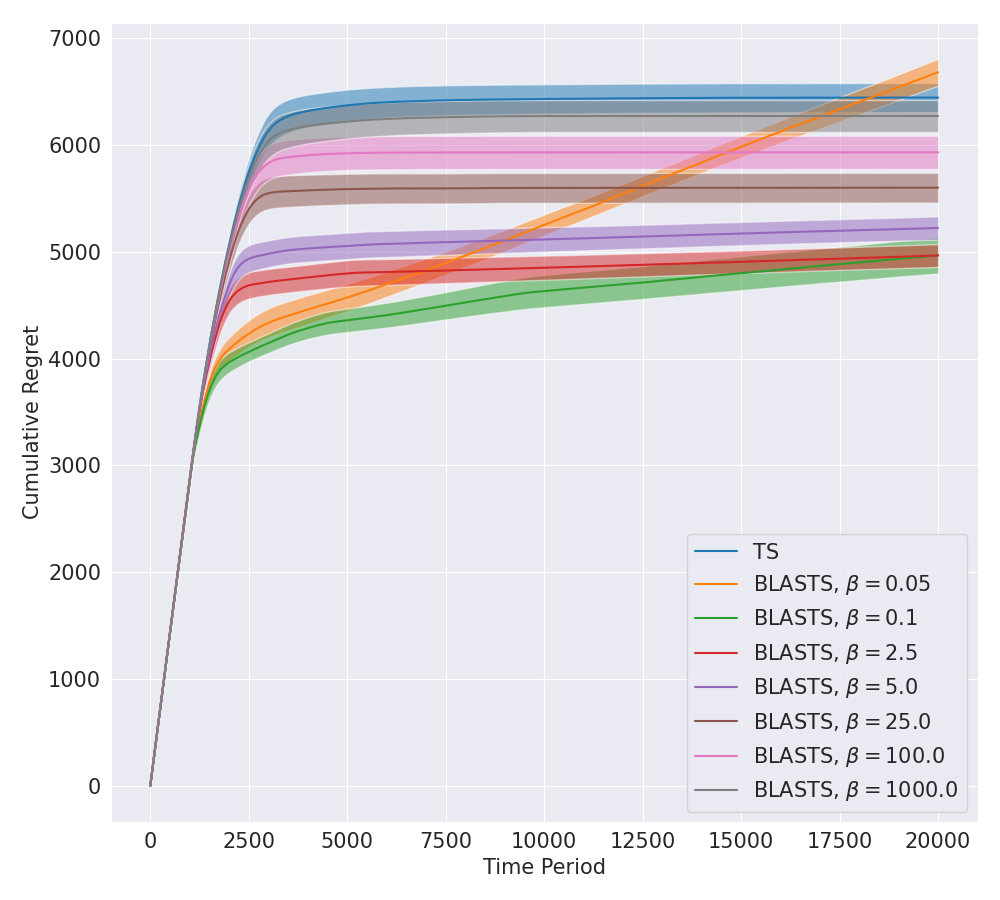}
  \caption{250 arms}
\end{subfigure}
\caption{Gaussian bandit with independent arms}
\label{fig:blasts_gauss}
\end{figure}

Our first experiment focuses on a Bernoulli bandit with $K$ independent arms. In each random trial, the environment is represented as a vector $\environment \in \bR^K$ where $\environment_a \sim \text{Uniform}(0,1), \forall a \in \mc{A}$. Accordingly, the reward observed for taking action $a \in \mc{A}$ is sampled as a $\text{Bernoulli}(\environment_a)$. In our second experiment, we pivot to a Gaussian bandit where rewards for action $a$ are drawn from $\mc{N}(\environment_a, 1)$, again with $\environment_a \sim \text{Uniform}(0,1), \forall a \in \mc{A}$. Results for each experiment are shown in Figures \ref{fig:blasts_bern} and \ref{fig:blasts_gauss}, respectively.

The first notable observation from both sets of experiments is the control that the $\beta$ parameter exerts over the performance of BLASTS. As expected, while $\beta \ra 0$, BLASTS approaches the performance of a uniform random policy. In contrast, as $\beta \ra \infty$, BLASTS gradually recovers the performance of Thompson sampling. Importantly, when obtaining a satisficing solution is viable, there is a suitable range of $\beta$ values to accommodate different degrees of sub-optimality, many of which converge to such satisficing policies in fewer time periods than what is needed for an optimal policy. In our experiments, we ran BLASTS for a wider range of $\beta$ values than what is shown and selectively pruned away a subset of values for readability. In all plots, the smallest value of $\beta$ in our selection that achieves the optimal policy is shown.

A second key finding of the above experiments is the capacity for BLASTS to synthesize an optimal policy more efficiently than Thompson sampling. Recall that the input $D$ to the rate-distortion function $\mc{R}(D)$ represents the desired upper bound on expected distortion. In the context of the Blahut-Arimoto algorithm, $\beta$ represents the desired slope of the recovered solution along the rate-distortion curve. By Corollary 5 of \citep{blahut1972computation}, we know that, given the current history $H_t$, the distortion $D$ achieved at the point on the rate-distortion curve parameterized by $\beta$ is given as $D(\beta | H_t) = \bE\left[\frac{\tilde{q}_\star(A) \exp(-\beta \E[(\overline{r}(A_\star) - \overline{r}(A))^2 | \environment, H_t])}{\sum_{a' \in \actions} \tilde{q}_\star(a') \exp(-\beta \E[(\overline{r}(A_\star) - \overline{r}(a'))^2 | \environment, H_t])}\right],$ where $\tilde{q}_\star$ achieves the infimum 
$$\inf\limits_{q} - \bE\left[\log\left(\sum\limits_{a \in \mc{A}} q(a)\exp(-\beta \E_t[(\overline{r}(A_\star) - \overline{r}(A))^2 | \environment]\right)\right].$$

Letting $\Delta > 0$ denote the action gap between the best and second-best arm~\citep{farahmand2011action,bellemare2016increasing}, it stands to reason that, for any $\beta$ obtaining the optimal policy, $\max\limits_t D(\beta | H_t) < \Delta^2$. By Fact \ref{fact:sa_less_info}, it follows that the target actions computed along these same $\beta$ values serve as easier learning targets (through smaller $\bI_t(\tilde{A};\environment)$) while still converging to the optimal policy.

In summary, the results presented here verify that BLASTS is capable of realizing a broad spectrum of policies. Included in this spectrum are satisficing policies that accommodate various problem constraints on time and resources, as well as optimal policies that be identified with greater efficiency than Thompson sampling.

\subsection{Balancing Rate-Distortion with the Information Ratio}

The previous experiments clearly illustrate the importance of the $\beta$ hyperparameter in enabling an agent designer to express preferences over behaviors and allowing an agent to realize those preferences through its learned target actions. In the context of the rate-distortion function, $\beta$ encodes a preference for minimizing rate over minimizing distortion. Some of the $\beta$ values that ultimately recover satisficing policies, however, do appear to have signs of strong performance in the earlier stages of learning. However, it is clear that despite this initial potential, the fixed value of $\beta$ is ultimately too small to prioritize regret minimization. It is a natural to wonder if allowing $\beta$ to vary with time might more efficiently synthesize an optimal policy? One crude strategy for exploring this would be to place $\beta$ on a manually-tuned schedule, eventually allowing it to increase to a value that emphasizes optimal actions by the end of learning. As a more principled alternative to such a laborious strategy, we consider the relationship between $\beta$ and the information ratio, inspired by the value of $\beta = \frac{1-\gamma^2}{(1-\gamma)^2 \Gamma}$ derived in our analysis.

\begin{figure}[H]
\centering
\begin{subfigure}{.5\textwidth}
  \centering
  \includegraphics[width=.75\linewidth]{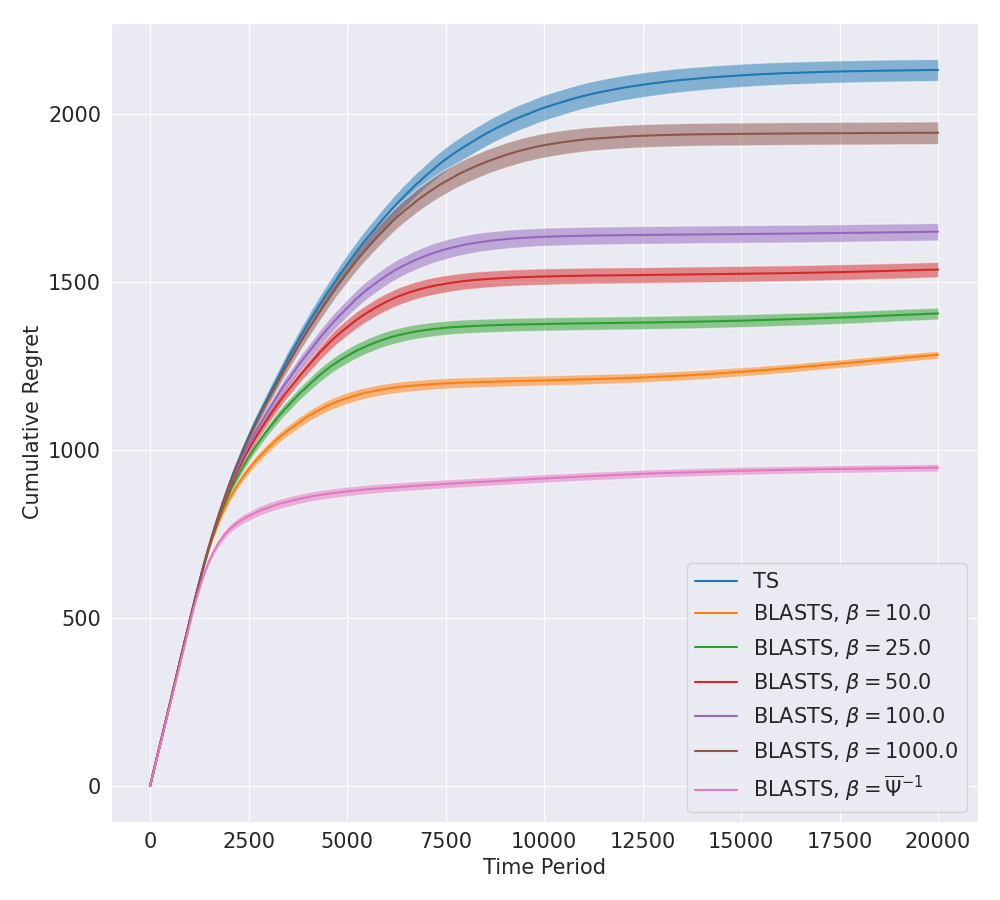}
  \caption{Bernoulli}
\end{subfigure}
\begin{subfigure}{.5\textwidth}
  \centering
  \includegraphics[width=.75\linewidth]{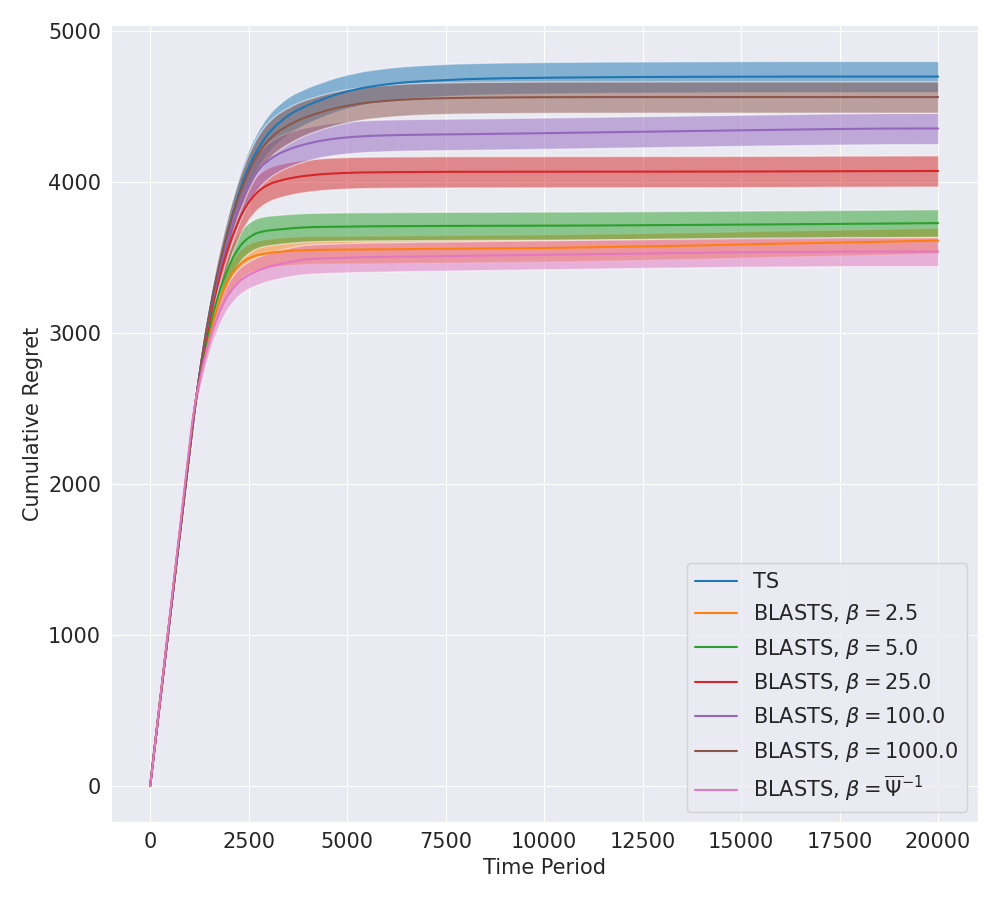}
  \caption{Gaussian}
\end{subfigure}
\caption{BLASTS with adaptive $\beta_t = \overline{\Psi}_t^{-1}$ for independent bandits with 50 arms}
\label{fig:blasts_info_ratio}
\end{figure}

The information ratio~\citep{russo2014learning,russo2016information,russo2018learning} is a powerful tool for expressing the cost (measured in squared units of regret) per bit of information acquired in each time period. The constant $\Gamma$ in our analysis acts a uniform upper bound on the information ratio (for our setting) that facilitates our information-theoretic regret bounds. For the more traditional setting of finding optimal policies, the information ratio at time period $t$ is given by $\Psi_t(\pi) = \frac{\Delta_t(\pi)^2}{g_t(\pi)}$ where $\Delta_t(\pi)$ denotes the expected regret with respect to $A_\star$ and $g_t(\pi)$ denotes the information gain $\bI_t(A_\star;A_t,O_{t+1})$. While, in theory, an agent wishes to compute a policy $\pi = \min\limits_\pi \Psi(\pi)$ that minimizes the information ratio, practical instantiations of this principle often rely on the fact that $g_t(\pi) \geq \bE[v_t(A)]$ where $v_t(A) = \bV[ \overline{r}(A) | \environment] | H_t]$ is the variance of the expected reward for action $A$ conditioned on the agent's current beliefs over the environment $\environment$~\citep{russo2014learning,russo2018learning}. Consequently, in each time period, an agent may aim to compute a policy that minimizes an upper bound $\overline{\Psi}(\pi) = \frac{\Delta_t(\pi)^2}{v_t(\pi)}.$
To see an initial connection between $\beta$ and the information ratio, recall that $\beta$ is representative of the desired slope along the rate-distortion curve~\citep{blahut1972computation}, with units of bits per unit of distortion; since BLASTS operates with a squared-regret distortion, this leaves $\beta$ as a quantity with units of bits per squared unit of regret. Moreover, once an agent has resolved most of its uncertainty in the environment, small values of the information ratio are indicative of optimal policies where BLASTS should, ideally, take on larger values of $\beta$ to identify such optimal actions. In light of these connections, we experiment with a version of BLASTS that uses the minimizer of the variance-based information ratio to compute $\beta$ in each time period. More specifically, let $\overline{\Psi}_t = \min\limits_{\pi \in \Delta(\mc{A})} \overline{\Psi}_t(\pi)$ and take $\beta_t = \overline{\Psi}_t^{-1}$; small constant is always added to $\overline{\Psi}_t$ to avoid division by zero. Results for this variant on the independent Bernoulli and Gaussian bandits are shown in Figure \ref{fig:blasts_info_ratio}. While an adaptive $\beta$ shows marginal gain in the Gaussian bandit, the Bernoulli bandit results show marked improvement in finding an optimal policy.

These results using an adaptive $\beta_t$ can be translated back to the fixed $\beta$ setting by considering a distortion function $\hat{d}(\tilde{a},e) = \overline{\Psi}^{-1} d(\tilde{a},\environment)$. Our choice of using expected squared distortion is supported by our theory, however the question of whether more efficient distortion functions exist in practice is an interesting direction for future work.

\section{Conclusion}
\label{sec:conc}

A standard design principle of sequential decision-making is to build agents that learn optimal actions. Recent work has highlighted scenarios wherein problem constraints make the pursuit of optimal actions infeasible, forcing the agent designer to craft a new target for an agent to learn. In this work, we forge a new direction where agents are designed to fabricate their own learning targets whose generic form is now the sole responsibility of the agent designer. We highlight how rate-distortion theory gives rise to a principled form for these learning targets, allowing practitioners to express their preference between the ease of learning and the sub-optimality of the resulting policy. We prove a general regret bound for this setting, contending with the non-stationarity of learning targets, and empirically verify the flexibility of our approach in yielding a broad spectrum of policies with varying degrees of sub-optimality. Importantly, we find that an agent's ability to specify target actions that require fewer bits of information can translate into greater efficiency in finding optimal policies relative to Thompson sampling. Future work may find it fruitful to couple the Blahut-Arimoto algorithm with more powerful strategies for information acquisition~\citep{russo2018learning}.

\section*{Acknowledgements}

Financial support from Army Research Office (ARO) grant W911NF2010055 is gratefully acknowledged.

\bibliographystyle{icml2021}
\bibliography{references}

\appendix

\onecolumn

\section{Related Work}
\label{sec:related}

Our work focuses on principled Bayesian exploration wherein an agent maintains a posterior distribution over its environment~\citep{chapelle2011empirical,agrawal2012analysis,agrawal2013further,russo2016information}. As complete knowledge of the environment (the vector of mean rewards at each arm, for example) would endow an agent with prescience of optimal actions, efficient exploration amounts to the resolution of an agent's epistemic uncertainty about the environment. A natural approach for resolving such uncertainty is Thompson sampling which employs probability matching in each time period to sample actions according to the probability of being optimal~\citep{thompson1933likelihood,agrawal2012analysis,agrawal2013further,russo2016information,russo2018tutorial}. \citet{chapelle2011empirical} kickstarted renewed interest in Thompson sampling through empirical successes in online advertisement and news recommendation applications. While a corresponding regret bound was developed in subsequent work~\citep{agrawal2012analysis,agrawal2013further}, our paper follows suit with \citet{russo2016information} who introduced an elegant, information-theoretic analysis of Thompson sampling; their technique has been subsequently studied and extended to a variety of other problem settings~\citep{russo2018learning,russo2018satisficing,dong2018information} and applications~\citep{lattimore2019information,osband2019deep}. In this work, we also leverage the information-theoretic analysis of \citet{russo2016information} while additionally incorporating ideas from rate-distortion theory~\citep{shannon1959coding}. Unlike prior work exploring the intersection of sequential decision-making and rate-distortion theory, we are not concerned with state abstraction~\citep{abel2019state} nor are we concerned with an agent exclusively targeting optimal actions through some compressive statistic of the environment~\citep{dong2018information}.

A core novelty of this paper is leveraging the Blahut-Arimoto algorithm~\citep{arimoto1972algorithm,blahut1972computation} for the efficient computation of rate-distortion functions. The algorithm was originally developed for the dual problem of computing the channel-capacity function~\citep{arimoto1972algorithm} and was soon after extended to handle computation of the rate-distortion function as well~\citep{blahut1972computation}. An initial study of the algorithm's global convergence properties (for discrete random variables) was done by \citet{arimoto1972algorithm} and further explored by \citet{csiszar1974computation,csiszar1984information}. While there have been many variants of the Blahut-Arimoto algorithm introduced over the years~\citep{sayir2000iterating,matz2004information,vontobel2008generalization,naja2009geometrical,yu2010squeezing}, we find that the simplicity of the original algorithm is suitable both in theory and in practice. 

The goal of finding target actions with a tolerable degree of sub-optimality deviates from the more traditional objective of identifying optimal actions. As previously mentioned, this setting can implicitly arise when faced with a continuous action space~\citep{bubeck2011x,kleinberg2008multi,rusmevichientong2010linearly}, a fixed time horizon~\citep{ryzhov2012knowledge,deshpande2012linear}, or an infinite-armed bandit problem~\citep{berry1997,Wang2008AlgorithmsFI,bonald2013two}. \citet{russo2018satisficing} attempt to rectify some shortcomings of these works by introducing a discounted notion of regret that emphasizes initial stages of learning and measures performance shortfall relative to satisficing actions, instead of optimal ones. Moreover, the analysis of their satisficing Thompson sampling algorithm inherits the benefits of flexibility and generality from the analogous information-theoretic results for Thompson sampling~\citep{russo2016information}. In this work, we obviate the need for the manual specification of satisficing actions, instead relying on direct computation of the rate-distortion function to adaptively compute the distribution over satisficing actions in each time period that achieves the rate-distortion limit.

The idea of an agent that learns to designate and achieve its own goals bears close resemblance to hierarchical agents studied in the reinforcement-learning literature~\citep{kaelbling1993hierarchical,dayan1993feudal,sutton1999between,barto2003recent}. In recent years, the two most-popular paradigms for hierarchical reinforcement learning have been feudal reinforcement learning~\citep{dayan1993feudal,nachum2018data} and options~\citep{sutton1999between,jong2008utility,bacon2017option,wen2020efficiency}. Feudal reinforcement-learning agents are comprised of an internal managerial hierarchy wherein the action space of managers represents sub-goals for workers in the subsequent level of the hierarchy; when workers can be quickly trained to follow the directed sub-goals of their managers (without regard for the optimality of doing so) the top-most manager can more efficiently synthesize an optimal policy. Options provide a coherent abstraction for expressing various temporally-extended behaviors or skills, typically replacing or augmenting the original action space of the agent~\citep{jong2008utility}. While there is great empirical support for the performance of feudal learning and options when the goal representation or option set is computed and fixed a priori, recent work in learning such components online often relies on laborious tuning and heuristics to achieve success~\citep{vezhnevets2017feudal,bacon2017option,harb2018waiting}. In contrast, the main contribution of this work is to build a principled approach for learning such targets, albeit with a restricted focus to the simpler setting of bandit learning. We leave the exciting question of how the ideas presented here may scale up to tackle the challenges of hierarchical reinforcement learning to future work.

\section{Blahut-Arimoto Satisficing Thompson Sampling}
\label{sec:blasts_alg}

Here we present the full BLASTS algorithm with inline comments for clarity.

\input{algorithms/blasts_alg}

\section{Discounted Regret Analysis}
\label{sec:regret_proofs}

Abstracting away the precise details of BLASTS, we can consider a coarsely-defined algorithm that selects each action $A_t$ as follows: \textbf{(1)} identify a target action $\tilde{A}_t$ that minimizes a loss function $\mathcal{L}_\beta(\cdot|H_t)$ and \textbf{(2)} sample $A_t \sim \Pr(\tilde{A}_t = \cdot | H_t)$. Recall that the loss function is defined, for any target action $\tilde{A}$, by
$$\mathcal{L}_\beta(\tilde{A}|H_t) = \mathbb{I}_t(\environment; \tilde{A}) +  \beta \E_t\left[(\overline{r}(A_\star) - \overline{r}(\tilde{A}))^2 \right].$$
The following result helps establish that the expected loss of any target action decreases as observations accumulate.
\begin{lemma}
For all $\beta > 0$, target actions $\tilde{A}$, and $t = 0,1,2,\ldots$, 
$$\E_t[\mathcal{L}_\beta(\tilde{A} | H_{t+1})] = \mathcal{L}_\beta(\tilde{A} | H_t) - \mathbb{I}_t(\tilde{A};(A_t, O_{t+1})).$$
\end{lemma}
\input{proofs/lemma1}
As a consequence of the above, the following lemma assures that expected loss decreases as target actions are adapted.  It also suggests that there are two sources of decrease in loss: (1) a possible decrease in shifting from target $\tilde{A}_t$ to $\tilde{A}_{t+1}$ and (2) a decrease of $\mathbb{I}_t(\tilde{A}_t; (A_t, O_{t+1}))$ from observing the interaction $(A_t, O_{t+1})$.  The former reflects the agent's improved ability to select a suitable target, and the latter captures information gained about the previous target.  We omit the proof as the lemma follows immediately from Lemma \ref{le:information-supermartingale} and the fact that $\tilde{A}_{t+1}$ minimizes $\mathcal{L}_\beta(\tilde{A}_{t+1} | H_{t+1})$, by definition.
\begin{lemma}
For all $\beta > 0$, target actions $\tilde{A}$, and $t = 0,1,2,\ldots$, 
$$\E[\mathcal{L}_\beta(\tilde{A}_{t+1} | H_{t+1}) | H_t] \leq \E[\mathcal{L}_\beta(\tilde{A}_t | H_{t+1}) | H_t] = \mathcal{L}_\beta(\tilde{A}_t | H_t) - \mathbb{I}_t(\tilde{A}_t;(A_t, O_{t+1})).$$
\end{lemma}

Note that, for all $t$, loss is non-negative and bounded by mutual information between the optimal action and the environment (since optimal actions incur a distortion of 0):
$$\mathcal{L}_\beta(\tilde{A}_t | H_t) \leq \mathcal{L}_\beta(A_\star | H_t) = \mathbb{I}_t(\environment; A_\star).$$
We therefore have the following corollary.
\begin{corollary}
For all $\beta > 0$ and $\tau = 0,1,2,\ldots$, 
$$\E\left[\sum_{t=\tau}^\infty \mathbb{I}_t(\tilde{A}_t; (A_t, O_{t+1})) \Big| H_\tau\right] 
\leq \mathbb{I}_\tau(\environment; A_\star).$$
\end{corollary}
We omit the proof of Corollary 1 as it follows directly by applying the preceding inequality to the following  generalization that applies to any target action.
\begin{corollary}
For all $\beta > 0$, target actions $\tilde{A}$, and $\tau = 0,1,2,\ldots$, 
$$\E_\tau\left[\sum_{t=\tau}^\infty \mathbb{I}_t(\tilde{A}_t; (A_t, O_{t+1})) \right] 
\leq \mathcal{L}_\beta(\tilde{A} | H_\tau).$$
\end{corollary}
\input{proofs/corollary2}

Let $\Gamma$ be a constant such that 
$$\Gamma \geq \frac{\E_t[\overline{r}(\tilde{A}) - \overline{r}(A)]^2}{\mathbb{I}_t(\tilde{A}; A, O)},$$
for all histories $H_t$, target actions $\tilde{A}$, if the executed action $A$ is an independent sample drawn from the marginal distribution of $\tilde{A}$, and $O$ is the resulting observation. Thus, $\Gamma$ is an upper bound on the information ratio~\citep{russo2014learning,russo2016information,russo2018learning} for which existing information-theoretic analyses of worst-case finite-arm bandits and linear bandits provide explicit values of $\Gamma$ that satisfy this condition.

We can now establish our main results.  We omit the proof of Theorem \ref{thm:rb_optimact} as it is a special case of our subsequent result.
\begin{theorem}
If $\beta = \frac{1-\gamma^2}{(1-\gamma)^2 \Gamma}$ then, for all $\tau = 0,1,2,\ldots$,
$$\E_\tau\left[\sum_{t=\tau}^\infty \gamma^{t-\tau} (\overline{r}(A_\star) - \overline{r}(A_t)) \right] \leq 2 \sqrt{\frac{\Gamma \mathbb{I}_\tau(\environment; A_\star)}{1-\gamma^2}}.$$
\end{theorem}

In a complex environment with many actions, $\mathbb{I}(\environment; A_\star)$ can be extremely large, rendering the above result somewhat vacuous under such circumstances.  The next result offers a generalization, establishing a regret bound that can depend on the information content of any target action, including of course those that are much simpler than $A_\star$.
\begin{theorem}
If $\beta = \frac{1-\gamma^2}{(1-\gamma)^2 \Gamma}$ then, for all target actions $\tilde{A}$ and $\tau = 0,1,2,\ldots$,
$$\E_\tau\left[\sum_{t=\tau}^\infty \gamma^{t-\tau} (\overline{r}(A_\star) - \overline{r}(A_t))\right] \leq 2 \sqrt{\frac{\Gamma \mathbb{I}(\environment; \tilde{A} |H_\tau = H_\tau)}{1-\gamma^2}} + \frac{2\epsilon}{1-\gamma},$$
where $\epsilon = \sqrt{\E[(\overline{r}(A_\star) - \overline{r}(\tilde{A})^2 | H_\tau]}$.
\end{theorem}
\input{proofs/theorem2}

\section{Undiscounted Regret Analysis}

In this section, we derive a variant of Theorem 2 where performance shortfall is measured by the expected cumulative regret across a finite horizon. Consider a fixed time horizon $T$ and observe the analogous result to Corollary 2:
\begin{corollary}
For all $\beta > 0$, target actions $\tilde{A}$, and $\tau = 0,1,2,\ldots$, 
$$\E_\tau\left[\sum_{t=\tau}^{T + \tau} \mathbb{I}_t(\tilde{A}_t; (A_t, O_{t+1})) \right] 
\leq \mathcal{L}_\beta(\tilde{A} | H_\tau).$$
\end{corollary}
\input{proofs/corollary3}

With Corollary \ref{co:cumulative-information_undisc}, we may introduce the undiscounted analog to Theorem \ref{thm:rb_targetact}:

\begin{theorem}
If $\beta = \frac{T}{\Gamma}$ then, for all target actions $\tilde{A}$ and $\tau = 0,1,2,\ldots$,
$$\E_\tau\left[\sum_{t=\tau}^{T + \tau} \overline{r}(A_\star) - \overline{r}(A_t)\right] \leq 2 \sqrt{\Gamma T \mathbb{I}_\tau(\environment; \tilde{A})} + 2T\epsilon,$$
where $\epsilon = \sqrt{\E[(\overline{r}(A_\star) - \overline{r}(\tilde{A})^2 | H_\tau]}$.
\end{theorem}
\input{proofs/theorem3}

\end{document}

%% file: commands.tex
\definecolor{dblue}{RGB}{98, 140, 190}
\definecolor{dlblue}{RGB}{216, 235, 255}
\definecolor{dgreen}{RGB}{124, 155, 127}
\definecolor{dpink}{RGB}{207, 166, 208}
\definecolor{dyellow}{RGB}{255, 248, 199}
\definecolor{dgray}{RGB}{46, 49, 49}

\newcommand{\durl}[1]{\textcolor{dblue}{\underline{\url{#1}}}}

\newcommand{\mc}[1]{\mathcal{#1}}

\newcommand{\bE}{\mathbb{E}}
\newcommand{\bV}{\mathbb{V}}
\newcommand{\bR}{\mathbb{R}}
\newcommand{\bP}{\mathbb{P}}
\newcommand{\bI}{\mathbb{I}}
\newcommand{\bH}{\mathbb{H}}
\newcommand{\bN}{\mathbb{N}}
\newcommand{\bF}{\mathbb{F}}

\newcommand{\expect}[1]{\mathbb{E}\left[#1\right]}

\newcommand{\ra}{\rightarrow}

\newmdenv[
  topline=false,
  bottomline=false,
  rightline = false,
  leftmargin=10pt,
  rightmargin=0pt,
  innertopmargin=0pt,
  innerbottommargin=0pt
]{innerproof}

\newenvironment{dproof}[1][Proof]{\begin{proof}[#1] \text{\vspace{2mm}} \begin{innerproof}}{\end{innerproof}\end{proof}\vspace{4mm}}

\newcounter{DaveDefCounter}
\newtheorem{corollary}{Corollary}

\newtheorem{lemma}{Lemma}

\newtheorem{theorem}{Theorem}
\newtheorem{fact}{Fact}

%% file: algorithms/blasts_alg_no_comments.tex
\begin{algorithm}[tbh]
   \caption{Blahut-Arimoto Satisficing Thompson Sampling (BLASTS)}
\begin{algorithmic}
   \STATE {\bfseries Input:} Lagrange multiplier $\beta \in \bR_{\geq 0}$, Blahut-Arimoto iterations $K \in \bN$, Posterior samples $Z \in \bN$
   \STATE $H_0 = \{\}$
   \FOR{$t=0$ {\bfseries to} $T-1$}
   \STATE $e_1,\ldots,e_Z \sim \bP(\environment \in \cdot | H_t)$
   \STATE $d(a,e|H_t) = \E[(\overline{r}(A_\star) - \overline{r}(a))^2 | \environment = e, H_t])$
   \STATE $\tilde{p}_0(a| e_z) = \frac{1}{|\mc{A}|}, \forall a \in \mc{A}, z \in [Z]$
   \FOR{$k=0$ {\bfseries to} $K-1$}
   \STATE $\tilde{q}_k(a) = \E_t[\tilde{p}_k(a|\environment)], \forall a \in \actions$
   \STATE $\tilde{p}_{k+1}(a |e_z) \propto \tilde{q}_k(a) \exp\left(-\beta d(a,e_z \mid H_t)\right), \forall a \in \actions, \forall z \in Z$ 
   \ENDFOR
   \STATE $\hat{z} \sim \text{Uniform}(Z)$
   \STATE $A_t \sim \tilde{p}_K(a|e_{\hat{z}})$
   \STATE $H_{t+1} = H_t \cup \{(A_t,O_{t+1})\}$
   \STATE $R_{t+1} = r(A_t, O_{t+1})$
   \ENDFOR
\end{algorithmic}
 \label{alg:blasts}
\end{algorithm}

%% file: algorithms/blasts_alg.tex
\begin{algorithm}[tbh]
   \caption{Blahut-Arimoto Satisficing Thompson Sampling (BLASTS)}
   \label{alg:blasts}
\begin{algorithmic}
   \STATE {\bfseries Input:} Lagrange multiplier $\beta \in \bR_{\geq 0}$, Blahut-Arimoto iterations $K \in \bN$, Posterior samples $Z \in \bN$
   \STATE $H_0 = \{\}$
   \FOR{$t=0$ {\bfseries to} $T-1$}
   \STATE $e_1,\ldots,e_Z \sim \bP(\environment \in \cdot | H_t)$ \COMMENT{Finite sample from current belief over $\environment$}
   \STATE $d(a,e|H_t) = \E[(\overline{r}(A_\star) - \overline{r}(a))^2 | \environment = e, H_t])$ \COMMENT{Distortion function for target action $\tilde{A}_t$}
   \STATE $\tilde{p}_0(a| e_z) = \frac{1}{|\mc{A}|}, \forall a \in \mc{A}, z \in [Z]$
   \FOR{$k=0$ {\bfseries to} $K-1$}
   \STATE $\tilde{q}_k(a) = \E_t[\tilde{p}_k(a|\environment)], \forall a \in \actions$ \COMMENT{Run the Blahut-Arimoto algorithm}
   \STATE $\tilde{p}_{k+1}(a |e_z) = \frac{\tilde{q}_k(a) \exp(-\beta d(a,e_z \mid H_t))}{\sum_{a' \in \actions} \tilde{q}_k(a') \exp(-\beta d(a,e_z \mid H_t))}, \forall a \in \actions, z \in [Z]$ 
   \ENDFOR
   \STATE $\hat{z} \sim \text{Uniform}(Z)$ \COMMENT{Select posterior sample uniformly at random}
   \STATE $A_t \sim \tilde{p}_K(a|e_{\hat{z}})$\COMMENT{Probability matching}
   \STATE $H_{t+1} = H_t \cup \{(A_t,O_{t+1})\}$
   \STATE $R_{t+1} = r(A_t, O_{t+1})$
   \ENDFOR
\end{algorithmic}
\end{algorithm}

%% file: proofs/lemma1.tex
\begin{dproof}
Recall that $H_{t+1} = (H_t, A_t, O_{t+1})$.  By definition of a target action, we have that $\forall t, H_t \perp \tilde{A} | \environment$, which implies $\bI_t((A_t,O_{t+1});\tilde{A}|\environment) = 0$. Thus, $$\bI_t(\environment; \tilde{A}) = \bI_t(\environment; \tilde{A}) + \bI_t((A_t,O_{t+1});\tilde{A}| \environment) = \mathbb{I}_t(\environment, (A_t, O_{t+1}); \tilde{A})$$ by the chain rule of mutual information. Applying the chain rule once again, we have,
$$\mathbb{I}_t(\environment; \tilde{A}) = \mathbb{I}_t(\environment, (A_t, O_{t+1}); \tilde{A}) = \mathbb{I}_t(\environment; \tilde{A} |A_t, O_{t+1}) + \mathbb{I}_t(\tilde{A}; (A_t, O_{t+1})).$$
It follows that
\begin{align*}
\E_t[\mathcal{L}_\beta(\tilde{A} | H_{t+1})] =& 
\E[\mathcal{L}_\beta(\tilde{A} | H_{t+1}) | H_t] \\
=& \E\left[\mathbb{I}_t(\environment; \tilde{A} |A_t, O_{t+1}) +  \beta \E\left[(\overline{r}(A_\star) - \overline{r}(\tilde{A}))^2 | H_t, A_t, O_{t+1} \right] \Big| H_t\right] \\
=& \E_t\left[\mathbb{I}_t(\environment; \tilde{A} | A_t, O_{t+1}) \right] +  \beta \E_t\left[(\overline{r}(A_\star) - \overline{r}(\tilde{A}))^2 \right] \\
=& \E_t\left[\mathbb{I}_t(\environment; \tilde{A}) - \mathbb{I}_t(\tilde{A};(A_t, O_{t+1})) \right]  +  \beta \E_t\left[(\overline{r}(A_\star) - \overline{r}(\tilde{A}))^2 \right] \\
=& \mathbb{I}_t(\environment; \tilde{A}) +  \beta \E_t\left[(\overline{r}(A_\star) - \overline{r}(\tilde{A}))^2 \right] - \mathbb{I}_t(\tilde{A};(A_t, O_{t+1})) \\
=& \mathcal{L}_\beta(\tilde{A} | H_t) - \mathbb{I}_t(\tilde{A};(A_t, O_{t+1})).
\end{align*}
\end{dproof}

%% file: proofs/corollary2.tex
\begin{dproof}
\begin{align*}
    \E_\tau\left[\sum_{t=\tau}^\infty \mathbb{I}_t(\tilde{A}_t; (A_t, O_{t+1}))\right] &\leq \E_\tau\left[\sum_{t=\tau}^\infty \mc{L}_\beta(\tilde{A}_t|H_t) - \bE_t\left[\mc{L}_\beta(\tilde{A}_{t+1}|H_{t+1})\right]\right] \\
    &= \sum_{t=\tau}^\infty \bE_\tau\left[\mc{L}_\beta(\tilde{A}_t|H_t) \right] - \bE_\tau\left[\bE_t\left[\mc{L}_\beta(\tilde{A}_{t+1}|H_{t+1})\right]\right] \\
    &= \bE_\tau\left[\mc{L}_\beta(\tilde{A}_\tau|H_\tau)\right] + \sum_{t=\tau+1}^\infty \bE_\tau\left[\mc{L}_\beta(\tilde{A}_t|H_t) \right] - \sum_{t=\tau}^\infty \bE_\tau\left[\mc{L}_\beta(\tilde{A}_{t+1}|H_{t+1}) \right] \\
    &= \mc{L}_\beta(\tilde{A}_\tau|H_\tau) + \sum_{t=\tau+1}^\infty \bE_\tau\left[\mc{L}_\beta(\tilde{A}_t|H_t) \right] - \sum_{t=\tau+1}^\infty \bE_\tau\left[\mc{L}_\beta(\tilde{A}_{t}|H_{t}) \right] \\
    &= \mc{L}_\beta(\tilde{A}_\tau|H_\tau) \leq \mc{L}_\beta(\tilde{A}|H_\tau)
\end{align*}

where the steps follow as Lemma \ref{le:adapt_targets}, linearity of expectation, the tower property, and the fact that $\tilde{A}_\tau$ is the minimizer of $\mc{L}_\beta(\cdot|H_\tau)$, by definition.
\end{dproof}

%% file: proofs/theorem2.tex
\begin{dproof}
From the inequalities satisfied by $\Gamma$, the Cauchy-Schwartz inequality, and Corollary \ref{co:cumulative-information}, we have
\begin{align*}
\E_\tau\left[\sum_{t=\tau}^\infty \gamma^{t-\tau} (\overline{r}(\tilde{A}_t) - \overline{r}(A_t))\right]
\leq& \E_\tau\left[\sum_{t=\tau}^\infty \gamma^{t-\tau} \sqrt{\Gamma \mathbb{I}_\tau(\tilde{A}_t; (A_t, O_{t+1}))} \right] \\
\leq& \sum_{t=\tau}^\infty \sqrt{\gamma^{2 (t-\tau)} \Gamma} \sqrt{\sum_{t=\tau}^\infty \E_\tau\left[\mathbb{I}_\tau(\tilde{A}_t; (A_t, O_{t+1}))\right]} \\
\leq& \sqrt{\Gamma \mc{L}_\beta(\tilde{A} | H_\tau) \sum_{t=0}^\infty \gamma^{2t}} \\
=& \sqrt{\frac{\Gamma \mc{L}_\beta(\tilde{A} | H_\tau)}{1-\gamma^2}}.
\end{align*}

Since $\mathcal{L}_\beta(\tilde{A}_t | H_t) \geq 0$,
$$\sqrt{\E_t\left[(\overline{r}(A_\star) - \overline{r}(\tilde{A}_t))^2 \right]} 
\leq (1-\gamma) \sqrt{\frac{\Gamma \mathcal{L}_\beta(\tilde{A}_t | H_t)}{1-\gamma^2}}.$$
Further, applying Jensen's inequality to the left-hand side and using the fact that $\tilde{A}_t$ minimizes $\mathcal{L}_\beta(\tilde{A}_t | H_t)$ on the right-hand side, 
$$\E_t\left[\overline{r}(A_\star) - \overline{r}(\tilde{A}_t) \right] 
\leq (1-\gamma) \sqrt{\frac{\Gamma \mathcal{L}_\beta(\tilde{A} | H_t)}{1-\gamma^2}}.$$
Lemma \ref{le:information-supermartingale} implies that
$$\E_\tau[\mathcal{L}_\beta(\tilde{A} | H_t)] \leq \mathcal{L}_\beta(\tilde{A} | H_\tau),$$
for all $t \geq \tau$, and therefore, by Jensen's inequality,
$$\E_\tau\left[\overline{r}(A_\star) - \overline{r}(\tilde{A}_t) \right]
\leq (1-\gamma) \E_\tau\left[\sqrt{\frac{\Gamma \mathcal{L}_\beta(\tilde{A} | H_t)}{1-\gamma^2}} \right] 
\leq (1-\gamma) \sqrt{\frac{\Gamma \E_\tau\left[\mathcal{L}_\beta(\tilde{A} | H_t) \right]}{1-\gamma^2}} 
\leq (1-\gamma) \sqrt{\frac{\Gamma \mathcal{L}_\beta(\tilde{A} | H_\tau)}{1-\gamma^2}}.$$
It follows that
\begin{align*}
\E_\tau\left[\sum_{t=\tau}^\infty \gamma^{t-\tau} (\overline{r}(A_\star) - \overline{r}(\tilde{A}_t)) \right] \leq& \sqrt{\frac{\Gamma \mathcal{L}_\beta(\tilde{A} | H_\tau)}{1-\gamma^2}} \\
\leq& \sqrt{\frac{\Gamma (\mathbb{I}_\tau(\environment; \tilde{A}) + \beta \epsilon^2)}{1-\gamma^2}} \\
\leq& \sqrt{\frac{\Gamma \mathbb{I}_\tau(\environment; \tilde{A}) }{1-\gamma^2}} + \frac{\epsilon}{1-\gamma}.
\end{align*}

Applying these same steps, we complete the above bound as $$ \E_\tau\left[\sum_{t=\tau}^\infty \gamma^{t-\tau} (\overline{r}(\tilde{A}_t) - \overline{r}(A_t))\right] \leq \sqrt{\frac{\Gamma \mathbb{I}_\tau(\environment; \tilde{A}) }{1-\gamma^2}} + \frac{\epsilon}{1-\gamma}.$$

Putting everything together, we have

\begin{align*}
    \E_\tau\left[\sum_{t=\tau}^\infty \gamma^{t-\tau} (\overline{r}(A_\star) - \overline{r}(A_t))\right] &= \E_\tau\left[\sum_{t=\tau}^\infty \gamma^{t-\tau} (\overline{r}(A_\star) - \overline{r}(\tilde{A}_t) + \overline{r}(\tilde{A}_t) - \overline{r}(A_t))\right] \\
    &= \E_\tau\left[\sum_{t=\tau}^\infty \gamma^{t-\tau} (\overline{r}(A_\star) - \overline{r}(\tilde{A}_t))\right] + \E_\tau\left[\sum_{t=\tau}^\infty \gamma^{t-\tau} (\overline{r}(\tilde{A}_t) - \overline{r}(A_t))\right] \\
    &\leq 2\sqrt{\frac{\Gamma \mathbb{I}_\tau(\environment; \tilde{A}) }{1-\gamma^2}} + \frac{2\epsilon}{1-\gamma}.
\end{align*}

\end{dproof}

%% file: proofs/corollary3.tex
\begin{dproof}
\begin{align*}
    \E_\tau\left[\sum_{t=\tau}^{T + \tau} \mathbb{I}_t(\tilde{A}_t; (A_t, O_{t+1}))\right] &\leq \E_\tau\left[\sum_{t=\tau}^{T + \tau} \mc{L}_\beta(\tilde{A}_t|H_t) - \bE_t\left[\mc{L}_\beta(\tilde{A}_{t+1}|H_{t+1})\right]\right] \\
    &= \sum_{t=\tau}^{T + \tau} \bE_\tau\left[\mc{L}_\beta(\tilde{A}_t|H_t) \right] - \bE_\tau\left[\bE_t\left[\mc{L}_\beta(\tilde{A}_{t+1}|H_{t+1})\right]\right] \\
    &= \bE_\tau\left[\mc{L}_\beta(\tilde{A}_\tau|H_\tau)\right] + \sum_{t=\tau+1}^{T + \tau} \bE_\tau\left[\mc{L}_\beta(\tilde{A}_t|H_t) \right] - \sum_{t=\tau}^{T + \tau} \bE_\tau\left[\mc{L}_\beta(\tilde{A}_{t+1}|H_{t+1}) \right] \\
    &= \mc{L}_\beta(\tilde{A}_\tau|H_\tau) + \sum_{t=\tau+1}^{T + \tau} \bE_\tau\left[\mc{L}_\beta(\tilde{A}_t|H_t) \right] - \sum_{t=\tau+1}^{T + \tau + 1} \bE_\tau\left[\mc{L}_\beta(\tilde{A}_{t}|H_{t}) \right] \\
    &= \mc{L}_\beta(\tilde{A}_\tau|H_\tau) - \bE_\tau\left[\mc{L}_\beta(\tilde{A}_{T + \tau +1}|H_{T + \tau +1}) \right] \\
    &\leq \mc{L}_\beta(\tilde{A}_\tau|H_\tau) \leq \mc{L}_\beta(\tilde{A}|H_\tau)
\end{align*}

where the steps follow as Lemma \ref{le:adapt_targets}, linearity of expectation, the tower property, the non-negativity of $\mc{L}_\beta(\tilde{A}_{t}|H_{t}) \geq 0$, and the fact that $\tilde{A}_\tau$ is the minimizer of $\mc{L}_\beta(\cdot|H_\tau)$, by definition.
\end{dproof}

%% file: proofs/theorem3.tex
\begin{dproof}
From the inequalities satisfied by $\Gamma$, the Cauchy-Schwartz inequality, and Corollary \ref{co:cumulative-information_undisc}, we have
\begin{align*}
\E_\tau\left[\sum_{t=\tau}^{T + \tau} \overline{r}(\tilde{A}_t) - \overline{r}(A_t)\right]
\leq& \sqrt{\Gamma} \E_\tau\left[\sum_{t=\tau}^{T + \tau} \sqrt{\mathbb{I}_\tau(\tilde{A}_t; (A_t, O_{t+1}))} \right] \\
\leq& \sqrt{\Gamma T \sum_{t=\tau}^{T + \tau} \E_\tau\left[\mathbb{I}_\tau(\tilde{A}_t; (A_t, O_{t+1}))\right]} \\
\leq& \sqrt{\Gamma T \mc{L}_\beta(\tilde{A} | H_\tau)}
\end{align*}

Since $\mathcal{L}_\beta(\tilde{A}_t | H_t) \geq 0$,
$$\sqrt{\E_t\left[(\overline{r}(A_\star) - \overline{r}(\tilde{A}_t))^2 \right]} 
\leq T^{-1} \sqrt{\Gamma T \mathcal{L}_\beta(\tilde{A}_t | H_t)}.$$
Further, applying Jensen's inequality to the left-hand side and using the fact that $\tilde{A}_t$ minimizes $\mathcal{L}_\beta(\tilde{A}_t | H_t)$ on the right-hand side, 
$$\E_t\left[\overline{r}(A_\star) - \overline{r}(\tilde{A}_t) \right] 
\leq T^{-1} \sqrt{\Gamma T \mathcal{L}_\beta(\tilde{A} | H_t)}.$$
Lemma \ref{le:information-supermartingale} implies that
$$\E_\tau[\mathcal{L}_\beta(\tilde{A} | H_t)] \leq \mathcal{L}_\beta(\tilde{A} | H_\tau),$$
for all $t \geq \tau$, and therefore, by Jensen's inequality,
$$\E_\tau\left[\overline{r}(A_\star) - \overline{r}(\tilde{A}_t) \right]
\leq T^{-1} \E_\tau\left[\sqrt{\Gamma T \mathcal{L}_\beta(\tilde{A} | H_t)} \right] 
\leq T^{-1} \sqrt{\Gamma T \E_\tau\left[\mathcal{L}_\beta(\tilde{A} | H_t) \right]} 
\leq T^{-1} \sqrt{\Gamma T \mathcal{L}_\beta(\tilde{A} | H_\tau)}.$$
It follows that
\begin{align*}
\E_\tau\left[\sum_{t=\tau}^{T + \tau} \overline{r}(A_\star) - \overline{r}(\tilde{A}_t) \right] \leq& \sqrt{\Gamma T \mathcal{L}_\beta(\tilde{A} | H_\tau)} \\
\leq& \sqrt{\Gamma T (\mathbb{I}_\tau(\environment; \tilde{A}) + \beta \epsilon^2)} \\
\leq& \sqrt{\Gamma T \mathbb{I}_\tau(\environment; \tilde{A})} + T\epsilon.
\end{align*}

Applying these same steps, we complete the above bound as $$ \E_\tau\left[\sum_{t=\tau}^{T + \tau} \overline{r}(\tilde{A}_t) - \overline{r}(A_t)\right] \leq \sqrt{\Gamma T \mathbb{I}_\tau(\environment; \tilde{A})} + T\epsilon.$$

Putting everything together, we have

\begin{align*}
    \E_\tau\left[\sum_{t=\tau}^{T + \tau} \overline{r}(A_\star) - \overline{r}(A_t)\right] &= \E_\tau\left[\sum_{t=\tau}^{T + \tau} \overline{r}(A_\star) - \overline{r}(\tilde{A}_t) + \overline{r}(\tilde{A}_t) - \overline{r}(A_t)\right] \\
    &= \E_\tau\left[\sum_{t=\tau}^{T + \tau} \overline{r}(A_\star) - \overline{r}(\tilde{A}_t)\right] + \E_\tau\left[\sum_{t=\tau}^{T + \tau} \overline{r}(\tilde{A}_t) - \overline{r}(A_t)\right] \\
    &\leq 2\sqrt{\Gamma T \mathbb{I}_\tau(\environment; \tilde{A})} + 2T\epsilon.
\end{align*}
\end{dproof}